\documentclass[journal]{IEEEtran}

\usepackage{epsfig} 
\usepackage{graphicx}
\usepackage{amsmath, amsfonts}
\usepackage[in]{fullpage}
\usepackage[small, bf]{caption}
\usepackage[normalem]{ulem}
\usepackage[margin=1in]{geometry}
\usepackage{floatrow}
\newfloatcommand{capbtabbox}{table}[][\FBwidth]
\usepackage[sorting=none]{biblatex}
\usepackage{booktabs}
\usepackage[utf8]{inputenc}
\usepackage{multirow} 
\usepackage{array}
\newcolumntype{C}[1]{>{\centering\let\newline\\\arraybackslash\hspace{0pt}}m{#1}}

\newcommand{\fig}[1]{Fig.~\ref{figure:#1}}

\newcommand{\figur}[1]{Figure~\ref{figure:#1}}

\newcommand{\eq}[1]{Eqn.~(\ref{equation:#1})}
\newcommand{\eqs}[2]{Eqns.~(\ref{equation:#1})~and~(\ref{equation:#2})}
\newcommand{\eqst}[3]{Eqns.~(\ref{equation:#1}),~(\ref{equation:#2})~and~(\ref{equation:#3})}
\newcommand{\eqsf}[4]{Eqns.~(\ref{equation:#1}),~(\ref{equation:#2}),~(\ref{equation:#3})~and~(\ref{equation:#4})}

\newcommand{\sect}[1]{Section~\ref{section:#1}}

\newcommand{\be}{\begin{equation}}
\newcommand{\ee}{\end{equation}}
\newcommand{\bea}{\begin{eqnarray}}
\newcommand{\eea}{\end{eqnarray}}
\newcommand{\beau}{\begin{eqnarray*}}
\newcommand{\eeau}{\end{eqnarray*}}
\newcommand{\bed}{\begin{displaymath}}
\newcommand{\eed}{\end{displaymath}}

\newcommand{\UT}{U_{\rm T}}

\newcommand{\Iout}{I_{\rm out}}
\newcommand{\Iin}{\ensuremath{I_{\rm in}}}
\newcommand{\Iunit}{\ensuremath{I_{\rm unit}}}

\begin{document}


\title{Analog Gated Recurrent Neural Network for Detecting Chewing Events}

\author{Kofi Odame, Maria Nyamukuru, Mohsen Shahghasemi, Shengjie Bi, David Kotz}

\maketitle

\begin{abstract}
We present a novel gated recurrent neural network to detect when a person is chewing on food. We implemented the neural network as a custom analog integrated circuit in a 0.18 $\mu$m CMOS technology. The neural network was trained on $6.4$ hours of data collected from a contact microphone that was mounted on volunteers' mastoid bones. When tested on $1.6$ hours of previously-unseen data, the neural network identified chewing events at a $24$-second time resolution. It achieved a recall of $91\%$ and an F1-score of $94\%$ while consuming $1.1~\mu$W of power. A system for detecting whole eating episodes---like meals and snacks---that is based on the novel analog neural network consumes an estimated $18.8~\mu$W of power.

\end{abstract}

\begin{IEEEkeywords}
Eating detection, wearable devices, analog LSTM, neural networks.
\end{IEEEkeywords}

\section{Introduction}

Monitoring food intake and eating habits are important for managing and understanding obesity, diabetes and eating disorders \cite{kang2017nutritional, o2014dietary, turton2018go}. Because self-reporting is unreliable, many wearable devices have been proposed to automatically monitor and record individuals' dietary habits \cite{bedri2017earbit, farooq2018accelerometer, bi2018auracle}. The challenge is that if these devices are too bulky (generally due to a large battery), or if they require frequent charging, then they intrude on the user's normal daily activities and are thus prone to poor user adherence and acceptance \cite{canhoto2017exploring, gao2015empirical, dunne2014social, hensel2006defining}.

We recently addressed this problem with a long short-term memory (LSTM) neural network for eating detection that is implementable on a low-power microcontroller \cite{nyamukuru2020tiny, amoh2019optimized}. However, our previous approach relied on a power-consumptive analog-to-digital converter (ADC). It also required the microcontroller unit (MCU) to unnecessarily spend power to process irrelevant (i.e. non-eating related) data.

\emph{Analog} LSTM neural networks have been proposed as a way to eliminate the ADC and also to minimize the microcontroller's processing of irrelevant data. Unfortunately, the state-of-the-art analog LSTMs \cite{jordan2020birhythmic, adam2018memristive, krestinskaya2018learning, han2019era, zhao2019long} are implemented with operational amplifiers (opamps), current/voltage converters, Hadamard multiplications and internal ADCs and digital-to-analog converters (DACs). These peripheral components represent a significant amount of overhead cost in terms of power consumption, which diminishes the benefits of an analog LSTM (see Table \ref{tab:overhead}). 

In this paper, we introduce a power-efficient analog neural network that contains no DACs, ADCs, opamps or Hadamard multiplications. Our novel approach is based on a current-mode adaptive filter, and it eliminates over $90\%$ of the power requirements of a more conventional solution.

\begin{figure}[t]
\begin{center}
\includegraphics[scale=0.6]{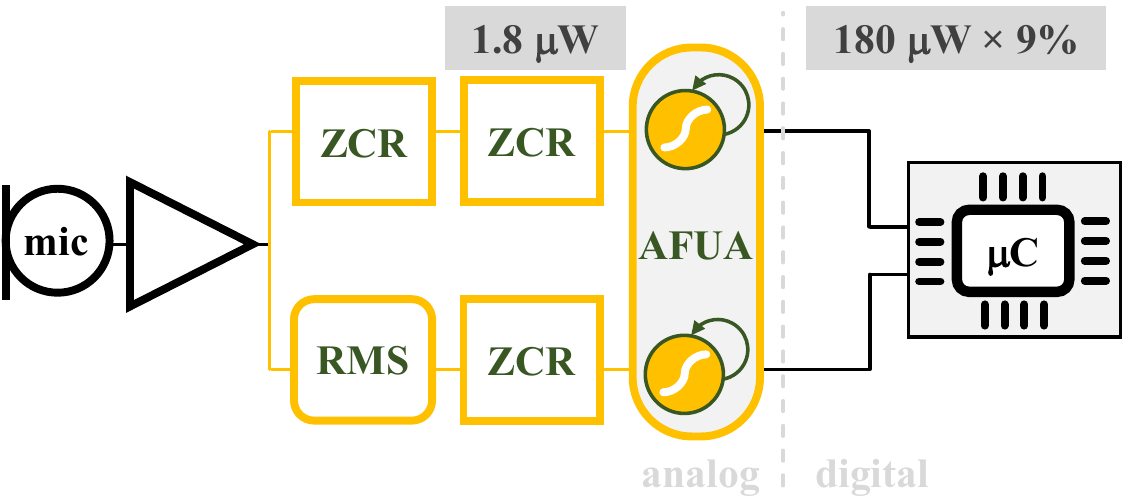}
\end{center}
\vspace{-10pt}
\caption{Block diagram of proposed eating detection system. From the contact microphone output, the ZCR and RMS blocks extract features based on zero-crossing rate and root-mean-square. The  analog neural network (labelled ``AFUA'') processes these features and produces a one-hot encoded output that predicts the presence or absence of a chewing event. The microcontroller (``$\mu$C'') merges and filters the individual chewing events into whole eating  episodes. The analog signal processing chain up to the AFUA block consumes 1.8$~\mu$W of power. The microcontroller is active only $9~\%$ of the time, during which it consumes $180$ $\mu$W of power.}
\label{figure:system}
\end{figure}







\section{Eating Detection System}
\figur{system} shows our proposed Adaptive Filter Unit for Analog (AFUA) long short-term memory as part of a signal processing system for detecting eating episodes. The input to the system is produced by a contact microphone that is mounted on the user's mastoid bone. Features are extracted from the contact microphone signal and input to the AFUA neural network, which infers whether or not the user is chewing. The AFUA's output is a one-hot encoding (($2, 0$)=chewing; ($0, 2$)=not chewing) of the predicted class label. Finally, a microcontroller processes the predicted class labels and groups the chewing events into discrete eating episodes, like a meal, or a snack \cite{bedri2017earbit, farooq2018accelerometer}. Following is a detailed description of the feature extraction and neural network components of the system.

\begin{table*}[t]
    \centering
    \begin{tabular}{|l|l|l|l|l|l|l|l|} \hline
    \multirow{2}{*}{\textbf{Circuit}} & \multirow{2}{*}{\textbf{Neuron Type}} & \multirow{2}{*}{\textbf{$m \times n$ }} & \multicolumn{5}{|c|}{\textbf{Power Consumption Overhead (\%)}  } \\ \cline{4-8}
       &&& \textbf {ADC} & \textbf{DAC} & \textbf{Buffer}  & \textbf{Opamp, V/I}   & \textbf{Total} \\ \hline \hline
    This work & AFUA & $10\times 16$ & 0 & 0 & 3 & 0 &  3 \\ \hline
    \cite{li2021ns} & GRU & $10\times 16$ & 0 & 0 &	32 & 0	&  32 \\ \hline
    \cite{han2019era} & LSTM & $128 \times 128$ & 12 & 25 &	1 & 30	&  68 \\ \hline
    \cite{zhao2019long} & LSTM & $16\times 16$ & 3 & 17 &	8 & 1	& 29 \\ \hline
    \end{tabular}
    \caption{Compared to other analog LSTM circuits, AFUA has the fewest peripheral components and hence the lowest overhead cost (see \sect{current_consumption}  derivation). $m$ and $n$ are number of hidden states and inputs, respectively. Note: for a given circuit, the larger the $m\times n$ product, the smaller the overhead. For fair comparison, we report AFUA overhead cost for $m \times n = 10 \times 16$.}
    \label{tab:overhead}
\end{table*}

\subsection{Feature Extraction}

\begin{figure}
\begin{center}
\includegraphics[scale=0.5]{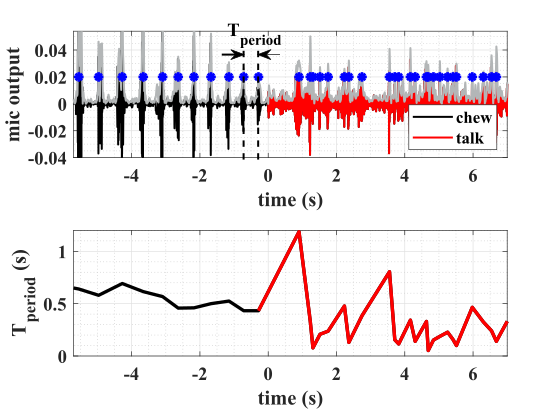}
\end{center}
\vspace{-20pt}
\caption{Typical time series data for chewing and talking events. Top panel: data from contact microphone shows that chewing (time $<0$ s) is characterized  by  quasi-periodic  bursts. No quasi-periodicity is observed during talking (time $\geq 0$ s). Bottom panel: duration between signal bursts (``T$_{\rm period}$''). For the chewing event (time $<0$ s), T$_{\rm period}$ is relatively constant. In contrast, T$_{\rm period}$ varies widely during the talking event.}
\label{figure:exampleData}
\end{figure}

As demonstrated in \fig{exampleData}, chewing is characterized by quasi-periodic bursts of large amplitude, low frequency  signals that can be measured by a contact microphone or accelerometer that is mounted on the head \cite{nyamukuru2020tiny, farooq2018accelerometer}. We can use the root mean square (RMS) and the zero-crossing rate (ZCR) to capture the signal's amplitude and frequency, respectively. A second ZCR operation applied to the RMS and the initial ZCR will produce information about the signal's periodicity. We implement the ZCR and RMS blocks based on the well-known rectifying current mirror. The details of the ZCR and RMS design may be found in \cite{baker2003micropower, sarpeshkar2005analog}. 

\subsection{AFUA Neural Network}
Fundamentally, an LSTM is a neuron that selectively retains, updates or erases its memory of input data \cite{hochreiter1997long}. The gated recurrent unit (GRU) is a simplified version of the classical LSTM, and it is described with the following set of equations \cite{cho2014learning}:
\begin{eqnarray}
r_j  & = & \sigma([\mathbf{W}_r \mathbf{x}]_j + [\mathbf{U}_r \mathbf{h}_{\langle t-1\rangle}]_j) \label{equation:r} \\ 
z_j  & = & \sigma([\mathbf{W}_z \mathbf{x}]_j + [\mathbf{U}_z \mathbf{h}_{\langle t-1\rangle}]_j) \label{equation:z} \\ 
\tilde{h}^{\langle t\rangle}_j & = & \tanh([\mathbf{W} \mathbf{x}]_j + [\mathbf{U}(\mathbf{r} \odot \mathbf{h}_{\langle t-1\rangle})]_j) \label{equation:htilde} \\ 
h^{\langle t\rangle}_j  & = & z_jh^{\langle t-1\rangle}_j + (1-z_j)\tilde{h}^{\langle t\rangle }_j, \label{equation:dh}
\end{eqnarray}
where $\mathbf{x}$ is the input, $h_j$ is the hidden state, $\tilde{h}_j$ is the candidate state, $r_j$ is the reset gate and $z_j$ is the update gate. Also, $\mathbf{W}_*$ and $\mathbf{U}_*$ are learnable weight matrices.

To implement the GRU in an efficient analog integrated circuit that contains no DACs, ADcs, operational amplifiers or multipliers, we can transform \eq{r}-(\ref{equation:dh}) as follows. The $\sigma$ function of \eq{z} gives $z_j$ a range of $(0, 1)$, and the extrema of this range reveals the basic mechanism of the update equation, \eq{dh}. For $z_j=0$, the update equation is $ h^{\langle t\rangle}_j  = \tilde{h}^{\langle t\rangle}_j$. For $z_j=1$, the update equation becomes $ h^{\langle t\rangle}_j  = h^{\langle t-1\rangle}_j$. Without loss of generality, we can replace $(1-z_j)$ with $z_j$ (this merely inverts the logic of the update gate, and inverts the sign of the $\mathbf{W}_z$ and $\mathbf{U}_z$ weight matrices). So, replacing $(1-z_j)$ and rearranging the update equation gives us
\begin{equation}
\left(h^{\langle t\rangle}_j-h^{\langle t-1\rangle}_j\right)/z_j + h^{\langle t-1\rangle}_j = \tilde{h}^{\langle t\rangle }_j,
\end{equation}
which is simply a first-order low pass filter with a continuous-time form of 
\begin{equation}
\frac{\tau}{z_j(t)}\frac{dh_j}{dt}+h_j(t) = \tilde{h}_j(t),
\label{equation:ct_update}
\end{equation}
where $\tau=\Delta T$, the time step of the discrete-time system. The gating mechanics of the continuous- versus discrete-time update equations are equivalent, modulo the inverted logic: For $z_j(t)=0$, \eq{ct_update} is a low-pass filter with an infinitely large time constant, and $h_j(t)$ does not change (this is equivalent to $ h^{\langle t\rangle}_j  = h^{\langle t-1\rangle}_j$ in discrete time). For $z_j(t)=1$, \eq{ct_update} is a low-pass filter with a time constant of $\tau=\Delta T$. Since the $\Delta T$ time step is small relative to the GRU's dynamics, a time constant of $\tau=\Delta T$ produces $h_j(t) \approx \tilde{h}_j(t)$ (equivalent to $ h^{\langle t\rangle}_j  = \tilde{h}^{\langle t\rangle}_j$ in discrete time).

Various studies have found the reset gate unnecessary with slow-changing signals, and also for event detection \cite{amoh2019optimized}. Both these scenarios describe our eating detection application, so we can discard the reset gate. 

Finally, if we translate the origins \cite{strogatz2018nonlinear} of both $h_j(t)$ and $\tilde{h}_j(t)$ to $1$, then we can replace the $tanh$ with a saturating function that has a range of $(0, 2)$. Such a saturating function can easily be implemented in analog circuitry, by taking advantage of the unidirectional nature of a transistor's drain-source current. We replace both the $tanh$ and the $\sigma$ with the following saturating function, 
\begin{equation}
f(y) = \frac{{\rm max}(y, 0)^2}{1 + {\rm max}(y, 0)^2},
\label{equation:activation}
\end{equation}
translate the origin and discard the reset gate to arrive at the \emph{Adaptive Filter Unit for Analog LSTM} (AFUA): 
\begin{eqnarray}
   z_j & = & f([\mathbf{W_{z} x}]_j + [\mathbf{U_{z} (h-1) + b_z}]_j) \label{equation:AFUAz} \\
   \tilde{h}_j & = & f([\mathbf{W x}]_j + [\mathbf{U (h-1) + b}]_j) \label{equation:AFUAh} \\
   \frac{\tau}{z_j}\frac{dh_j}{dt} & = & 2\tilde{h}_j - h_j,
   \label{equation:AFUAupdate}
\end{eqnarray}
where $[\cdot]_j$ is the j'th element of the vector. Also, $\mathbf{x}$ is the input, $h_j$ is the hidden state and $\tilde{h}_j$ is the candidate state. The variable $\tau$ is the nominal time constant, while $z_j$ controls the state update rate in \eq{AFUAupdate}. $\mathbf{W_{z}}$,  $\mathbf{U_{z}}$, $\mathbf{W}$,  $\mathbf{U}$ are learnable weight matrices, while $\mathbf{b_z}$, $\mathbf{b}$ are learnable bias vectors. The AFUA resembles the eGRU \cite{amoh2019optimized}, which we previously showed can be used for cough detection and keyword spotting. But while the eGRU is a conventional digital, discrete-time neural network, the AFUA is a continuous-time system, implementable as an analog integrated circuit.

\begin{figure}
\begin{center}
\includegraphics[scale=0.65]{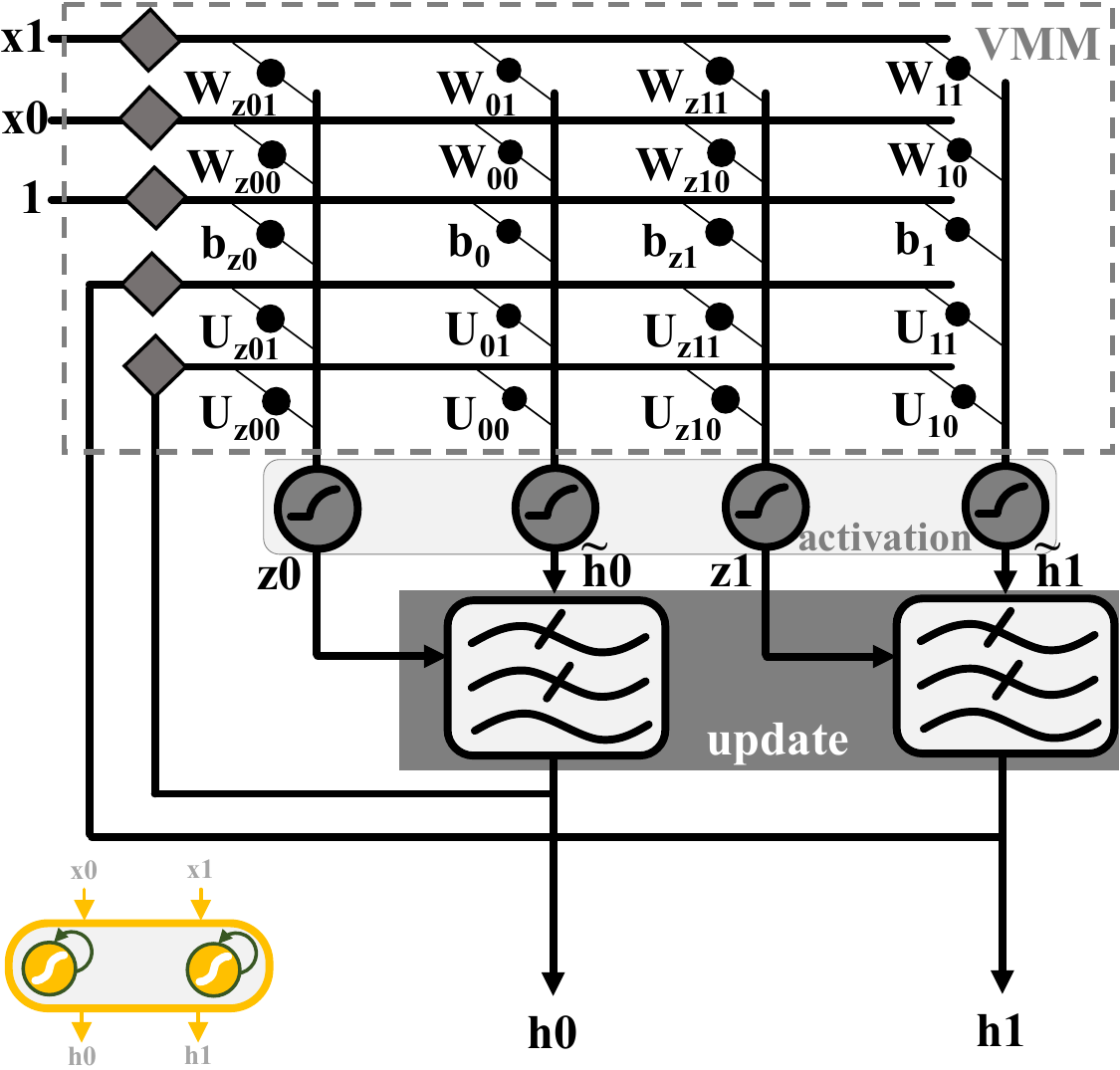}
\end{center}
\vspace{-10pt}
\caption{High level architecture of the AFUA neural network, which has a two-dimensional input feature vector, $\mathbf{x}=[x_0, x_1]^{\rm T}$. The network keeps a memory of past inputs by feeding back its  hidden states, $h_0, h_1$, to the vector matrix multiplier (VMM). The persistence of the network's memory depends on the time constants, $z_0, z_1$, of the adaptive low pass filters in the ``update'' block. Finally, the ``activation'' block provides saturating nonlinearities described by \eq{activation}.}
\label{figure:highLevel}
\end{figure}

\section{AFUA circuit implementation} 
\figur{highLevel} shows the high-level block diagram of the AFUA neural network. It comprises two AFUA cells (with corresponding hidden states $h_0$ and $h_1$), and it accepts two inputs, $x_0$ and $x_1$. Unlike previous LSTMs \cite{jordan2020birhythmic, adam2018memristive, krestinskaya2018learning, han2019era, zhao2019long}, the AFUA network contains no digital-to-analog converters, analog-to-digital converters, operational amplifiers or four-quadrant multipliers. Avoiding these power-consumptive components is what makes the AFUA implementation so efficient. Following are the circuit implementation details of the AFUA.

\subsection{Dimensionalization}
To realize the AFUA \eqsf{AFUAz}{AFUAh}{AFUAupdate}{activation} as an analog circuit, we first ``dimensionalize'' each variable and implement it as the ratio of a time-varying current and a fixed unit current, $\Iunit$ \cite{odame2005translinear, odame2009implementing}. For instance, we represent the update gate variable, $z_j$, as $I_z/\Iunit$.

\subsection{Activation Function}
\begin{figure}
\begin{center}
\includegraphics[scale=0.5]{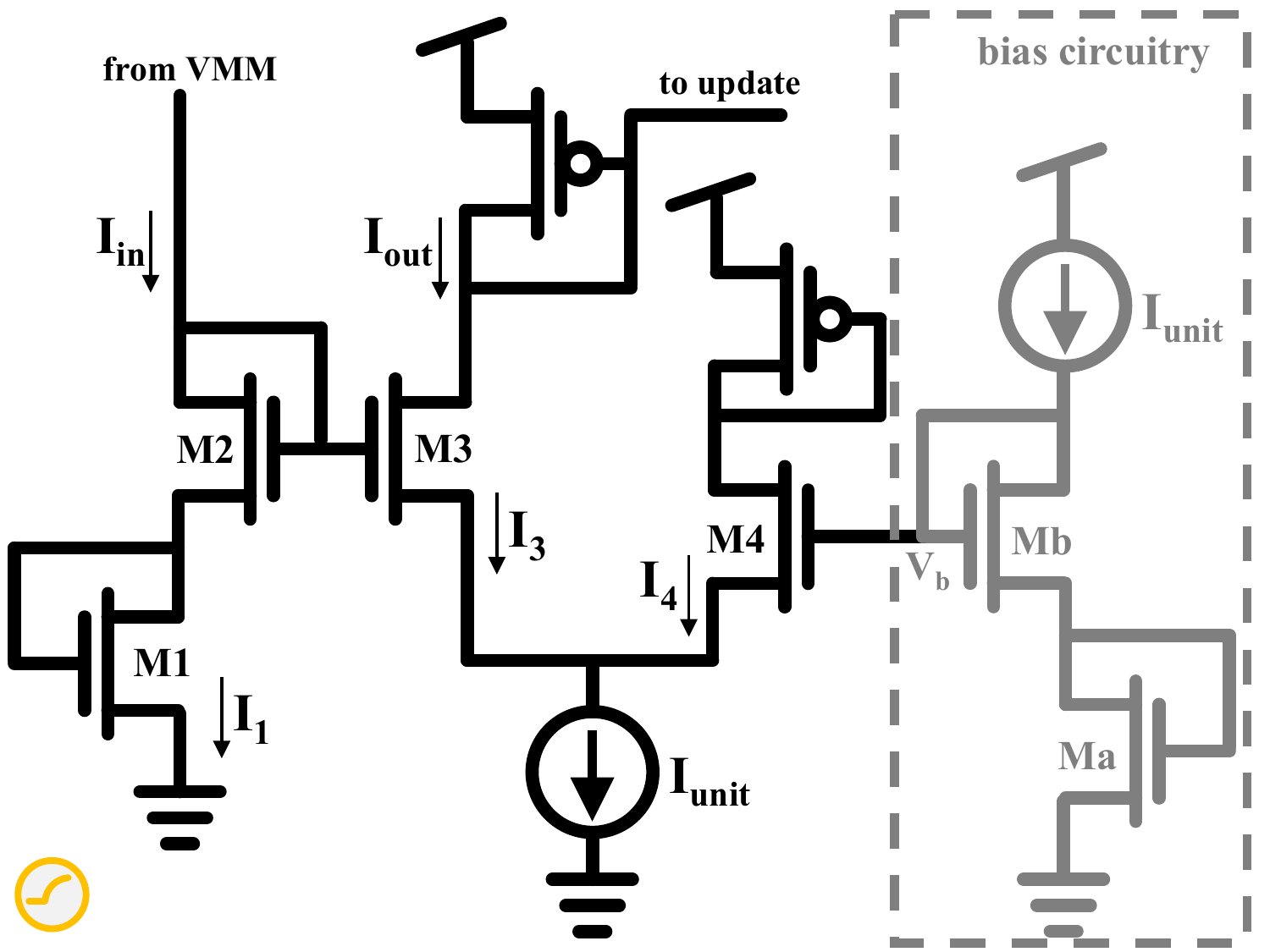}
\end{center}
\vspace{-10pt}
\caption{Activation function circuit schematic. A version of the input signal, $\Iin$, is reflected as current $\Iout$. The tail bias current source of the M$_3$-M$_4$ differential pair limits the output current to $\Iout < \Iunit$. Also, the one-sidedness of the nMOS drain current limits $\Iout$ to positive values only. In summary, the activation function circuit produces $0~\rm{A} \leq \Iout < \Iunit$.}
\label{figure:activSchem}
\end{figure}

\begin{figure}
\begin{center}
\includegraphics[scale=0.53]{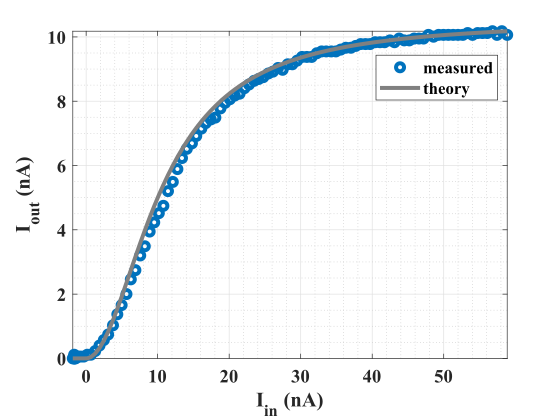}
\end{center}
\vspace{-10pt}
\caption{Activation function transfer curve. Chip measurements of the \fig{activSchem} circuit closely match the theoretically-predicted behavior of \eq{activation_ckt} for $\Iunit=10.5~$nA.}
\label{figure:activFunc}
\end{figure}

The \eq{activation} function is implemented as the current-starved current mirror shown in \fig{activSchem}. Kirchhoff's Current Law applied to the source of transistor M$_3$ gives
\begin{equation}
\Iout = I_3 = \Iunit - I_4.
\label{equation:activeKCL}
\end{equation}
The transistors are all sized equally, meaning that, from Kirchhoff's Voltage Law, the gate source voltage of transistor M$_3$ is
\begin{equation}
V_{\rm GS3} = 2V_{\rm GS1} + V_{\rm GS4} - 2V_{\rm GSa},
\label{equation:activeKVL}
\end{equation}
where we have assumed that the body effect in M$_2$ and M$_b$ is negligible.  If we operate the transistors in the subthreshold region, then \eq{activeKVL} implies
\begin{equation}
\Iout = I_3 = \frac{I_4 I_1^2}{\Iunit^2}.
\label{equation:activeTL}
\end{equation}
Combining \eqs{activeKCL}{activeTL} gives us
\begin{equation}
\Iout = \frac{\Iunit I_1^2}{\Iunit^2 +  I_1^2}.
\label{equation:active1}
\end{equation}
Now, the current flowing through a diode-connected nMOS is unidirectional, meaning $I_1 = {\rm max}(I_{\rm in}, 0)$, and we can write
\begin{equation}
\Iout = \Iunit\cdot\frac{{\rm max}(I_{\rm in}, 0)^2}{\Iunit^2 +  {\rm max}(I_{\rm in}, 0)^2},
\label{equation:activation_ckt}
\end{equation}
which is a dimensionalized analog of \eq{activation}. The measurement results in \fig{activFunc} illustrate the nonlinear, saturating behavior of this activation function.

\subsection{State Update}
\begin{figure}
\begin{center}
\includegraphics[scale=0.5]{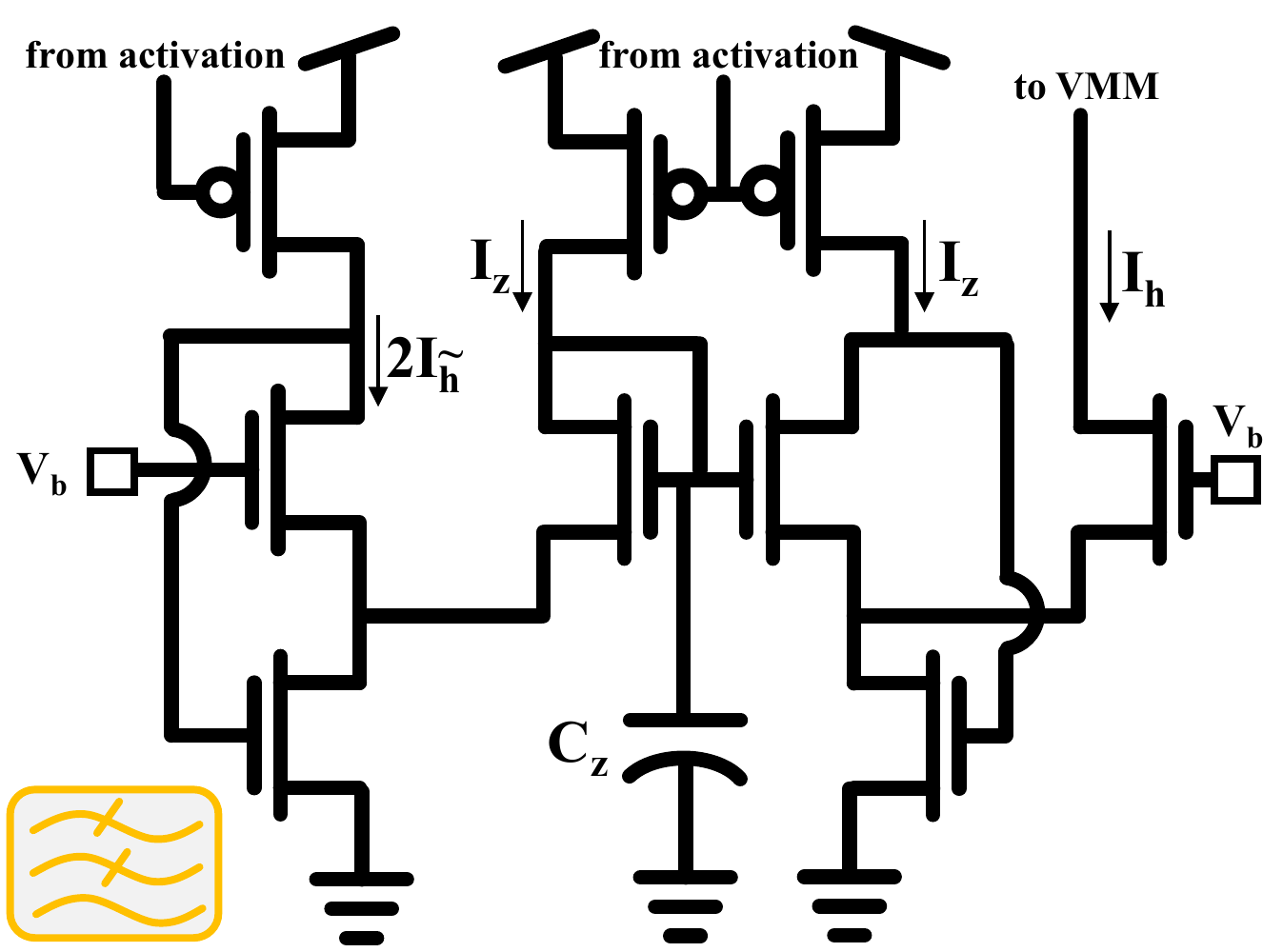}
\end{center}
\vspace{-10pt}
\caption{State update circuit schematic. The output $I_h$ is a low-pass-filtered version of the input, $2 I_{\tilde{h}}$. The filter's time constant is inversely proportional to the value of the current $I_z$. So, large values of $I_z$ increase the rate at which $I_h$ updates to $2 I_{\tilde{h}}$, while small values of $I_z$ slow down this process.}
\label{figure:updateSchem}
\end{figure}

\begin{figure}
\begin{center}
\includegraphics[scale=0.53]{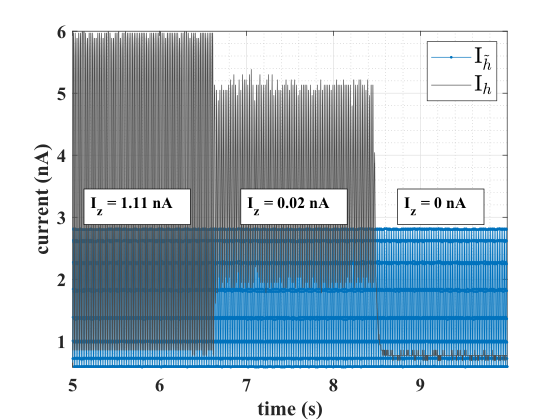}
\end{center}
\vspace{-10pt}
\caption{State update circuit response. Chip measurements of the \fig{updateSchem} circuit show that the output, $I_h$ follows the input, $I_{\tilde{h}}$ at a rate that is determined by the value of current $I_z$.}
\label{figure:updateResults}
\end{figure}

The AFUA state update, \eq{AFUAupdate}, is implemented as the adaptive filter shown in \fig{updateSchem}. The currents $I_h$, $I_{\tilde h}$ and $I_z$ represent the hidden state $h_j$, the candidate state $\tilde{h}_j$ and the update gate, $z_j$, respectively.  From the translinear loop principle, the \fig{updateSchem} circuit's dynamics can be written as \cite{mulder1996dynamic, odame2005translinear}
\begin{equation}
\underbrace{\frac{C_z \UT}{\kappa \Iunit}}_{\tau}\frac{\Iunit}{I_z}\frac{dI_h}{dt} = 2I_{\tilde h} - I_h,
\end{equation}
where $\kappa$ is the body-effect coefficient and $\UT$ is the thermal voltage \cite{EKV}. Just as $z_j$ does for $h_j$ in \eq{AFUAupdate}, $I_z$ controls the update speed of $I_h$ (see \fig{updateResults}).

\subsection{Vector Matrix Multiplication}
\begin{figure}
\begin{center}
\includegraphics[scale=0.85]{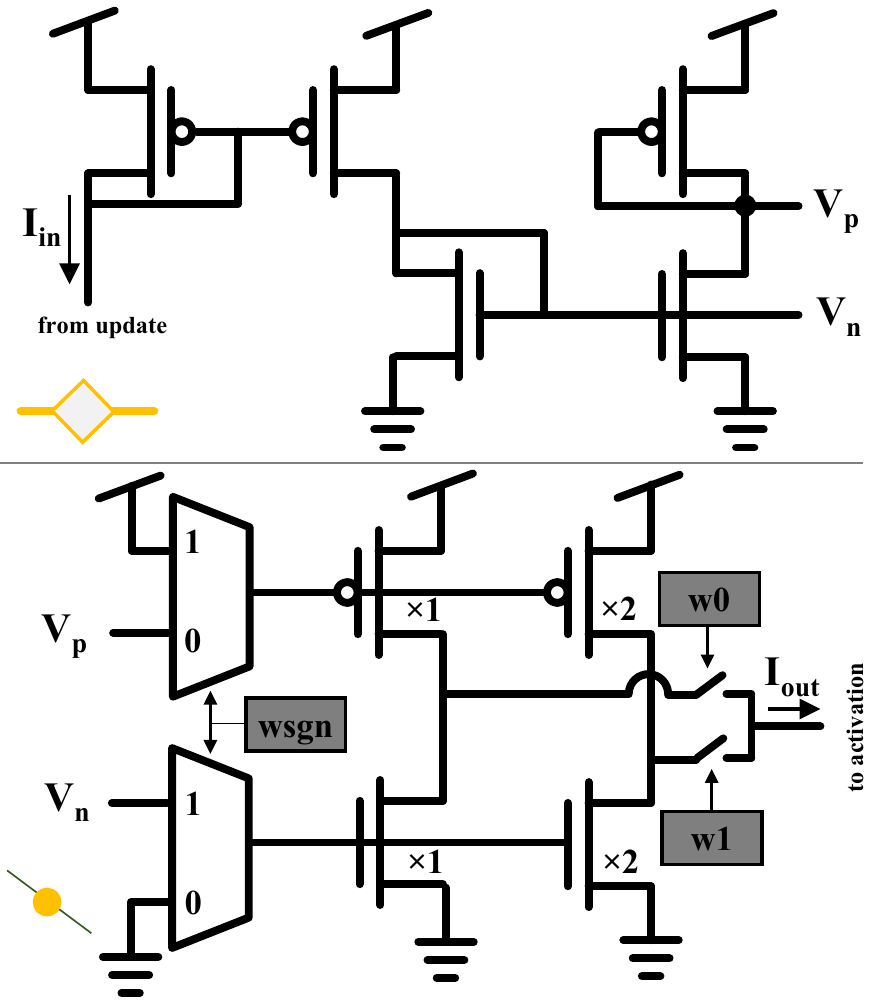}
\end{center}
\vspace{-10pt}
\caption{Vector matrix multiplier circuit components. The soma (top panel) and the synapse (bottom panel) form a programmable current mirror. The current mirror's gain is stored in registers w$_{\rm sgn}$, w$_0$, w$_1$. These represent the neural network's 3-bit quantized learned weights.}
\label{figure:somaCore}
\end{figure}

\figur{somaCore} depicts the components of our vector-matrix multiplication (VMM) block. These are the soma and synapse circuits that are common in the analog neuromorphic literature \cite{binas2016precise}. Crucially, the soma-synapse architecture is current-in, current-out. This means that, unlike other approaches for implementing GRU and LSTM networks \cite{adam2018memristive, krestinskaya2018learning, han2019era}, the VMM does not need power-consumptive operational amplifiers to convert signals between the current and voltage domains.

\section{Circuit Analysis}

\subsection{Current Consumption}
\label{section:current_consumption}
Since the activation function, \eq{activation}, has a range of $(0, 1)$, the $z_j$ and $\tilde{h}_j$ variables are likewise limited to $(0, 1)$. Also, from \eq{AFUAupdate}, $h_j$ spans $(0, 2)$. This means that all update gate and candidate state currents have a maximum value of \Iunit, while the hidden state currents have a maximum value of $2$\Iunit. With this information, we can calculate upper-bounds on the current consumption of each circuit component.

\begin{figure}
\begin{center}
\includegraphics[scale=0.53]{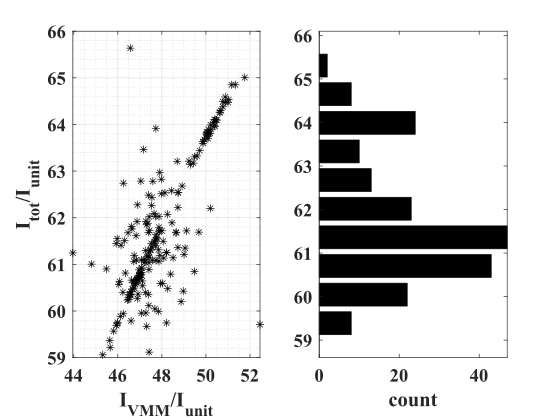}
\end{center}
\vspace{-10pt}
\caption{AFUA network current consumption when presented with 200 different input patterns. The scatter plot shows that the network's total current consumption is largely determined by the VMM. The average total current consumption is $62\Iunit$.}
\label{figure:powerDist}
\end{figure}

\subsubsection{Activation Function}
Not counting the input current that is supplied by the VMM, \fig{activSchem} shows that the only current consumed by the activation function block is the differential-pair tail current of \Iunit. There are two activation functions per AFUA cell (one each for $z_j$ and $\tilde{h}_j$). So, for an $m$-unit AFUA layer, the activation function blocks draw a total current of  $m\times 2\Iunit$.

\subsubsection{State Update}
The total current flowing through the four branches of the state update circuit (\fig{updateSchem}) is $2 \tilde{I}_h + 2 I_z +  I_h$, which has a worst-case value of $6\Iunit$. For our $m$-unit AFUA network, the state update circuits consume at most $m\times6 \Iunit$.

\subsubsection{VMM soma}
The soma is a current-mode buffer that drives a differential signal onto each row of the VMM (see \fig{somaCore}). For the somas on the input and bias rows, the maximum current consumption is $2$\Iunit. The somas driving the hidden state rows  consume at most $4$\Iunit~each. So, with $n$ inputs, $m$ hidden states and one bias row, the somas will consume a maximum total current of $(n+2m+1)\times2\Iunit$.

\subsubsection{VMM core}
As depicted in \fig{somaCore}, each multiplier element in the VMM core comprises a number of current sources that are switched on or off, depending on the values of the weight bits. At worst, all current sources are switched on, in which case the VMM elements that process state variables each consume $6\Iunit$, while those that process input variables or biases each consume $3\Iunit$. The maximum current draw of each VMM column for an $n$-input AFUA layer with $m$ hidden states is therefore $(n + 2m + 1)\times 3\Iunit$. There are $2m$ columns, to give a total maximum VMM core current consumption of $m(n + 2m + 1)\times 6\Iunit$.

\subsubsection{Total Current Consumption}

From the previous subsections, we conclude that the worst-case total current consumption of an $n$-input AFUA layer with $m$ hidden states is
\begin{equation}
I_{\rm tot} \leq (\underbrace{m(14+6(n+2m))}_{\rm core} + \underbrace{4m+2n+2}_{\rm VMM~soma})\times\Iunit,
\label{equation:Itot}
\end{equation}
where `core' includes the activation function, VMM core and state update current consumption. The VMM soma is peripheral to the AFUA's operation and represents overhead cost. For instance, a $16$-input, $10$-unit AFUA layer would spend $3~\%$ of its power budget as overhead.

Empirically, we found that the average current consumption of some of the AFUA blocks is significantly lower than their estimated worst-case values. In particular, the VMM consumes only $48\Iunit$ on average (see \fig{powerDist}). This leads to an average AFUA total current consumption of $62\Iunit$. The specific choice of $\Iunit$ depends on the desired operating speed, as we discuss in the following subsection.

\subsection{Estimated Power Efficiency}
\label{section:estimatedPower}
The power efficiency of neural networks is conventionally measured in operations per Watt. But this metric does not apply directly to a system like the AFUA, since it executes all of its operations continuously and simultaneously. However, we can estimate the AFUA's power efficiency by considering the performance of an equivalent discrete time system. 

To arrive at the discrete-form AFUA unit, we first replace the state variables of \eqst{AFUAz}{AFUAh}{AFUAupdate} with their discrete-time counterparts. This includes the discretization $dh_j/dt = (h^{\langle t \rangle}_j-h^{\langle t -1 \rangle}_j)/\Delta T$, where $\Delta T$ is the sampling period. Then, we set $\tau=\Delta T$ to produce the following expression.
\begin{eqnarray}
   z_j & = & f([\mathbf{W_{z} x}]_j + [\mathbf{U_{z} (h-1) + b_z}]_j) \nonumber \\
   \tilde{h}^{\langle t \rangle}_j & = & f([\mathbf{W x}]_j + [\mathbf{U (h}_{\langle t - 1 \rangle}\mathbf{-1) + b}]_j) \nonumber \\
   h^{\langle t \rangle}_j & = & z_j2\tilde{h}^{\langle t \rangle}_j - (1-z_j)h^{\langle t -1 \rangle}_j.
   \label{equation:AFUAupdateD}
\end{eqnarray}
Recall that $\mathbf{W}, \mathbf{W_{z}}$ are $2\times2$ matrices, $\mathbf{U}, \mathbf{U_{z}}$ are $1 \times 2$ vectors and $z_j$ are scalars, meaning that each discretized AFUA unit executes $14$ multiply operations per time step. Also, there are $2$ divisions due to the two activation functions (see \eq{activation}). Not counting additions and subtractions, each discretized AFUA unit executes $16$ operations per time step, to make for a total of $32$ operations/step performed by the network. Assuming the sampling period of $\Delta T=2$ ms used in our previous eating detection systems \cite{bi2018auracle, nyamukuru2020tiny}, this implies the AFUA performs the equivalent of $16,000$ operations per second.

Now, setting $\tau=\Delta T = 2$ ms requires a unit current of
\begin{equation}
\Iunit =  \frac{C_{\rm z} \UT}{\kappa \tau}  =  500\cdot \frac{C_{\rm z} \UT}{\kappa},
\label{equation:tauIunit}
\end{equation}
where $C_z=57$ fF is the integrating capacitor of the translinear loop filter, $\UT=26$ mV at room temperature and $\kappa \approx 0.42$. This gives $\Iunit = 1.8$ pA. With a total current consumption of $62 \Iunit$, a voltage supply of $1.8$ V and $16$K operations per second, the AFUA's equivalent operations per Watt is  $76$ TOps/W.


\subsection{Mismatch}
Due to random variations in doping and geometry, transistors that are nominally identical will exhibit mismatch when fabricated in a physical ASIC. To understand the effect of  mismatch and other non-idealities on the AFUA neural network's performance, we performed Monte Carlo analyses with foundry-provided manufacturing and test data. The Monte Carlo analyses included mismatch and process variation, as well as power supply voltage and temperature corners of $\{1.6 V, 2 V\}$ and $\{0  ^{\circ}C, 35 ^{\circ}C\}$, respectively. 

\begin{figure}
\begin{center}
\includegraphics[scale=0.54]{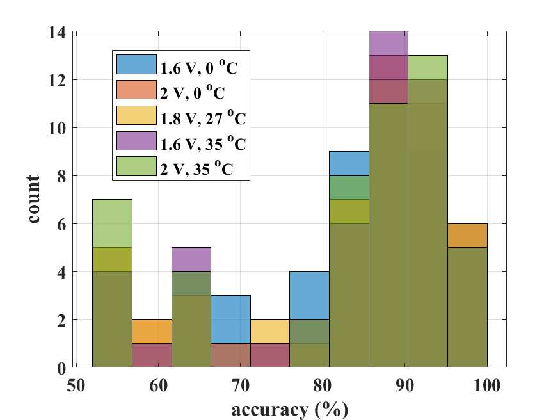}
\end{center}
\vspace{-10pt}
\caption{Monte Carlo analysis performed for $250$ runs, including mismatch and process variation, as well as power supply voltage and temperature corners of $\{1.6 V, 2 V\}$ and $\{0  ^{\circ}C, 35 ^{\circ}C\}$, respectively. Nominal power supply voltage and temperature are $1.8~V$, $27 ^{\circ}C$. Median accuracy is $90 ~\%$.}
\label{figure:monteCarlo}
\end{figure}

\figur{monteCarlo} shows the variation in classification accuracy for $250$ Monte Carlo runs of one implementation of the AFUA neural network. The median accuracy across all runs is $0.90$. Most of the variation in accuracy is due to mismatch, and the AFUA neural network is largely robust to temperature, voltage and process variation. The neural network is also unaffected by circuit noise (this is a direct result of the network's ability to generalize). To mitigate the effect of mismatch, we can use larger transistors \cite{pelgrom1989matching}, calibrate the network's learning algorithm for each individual chip \cite{binas2016precise}, or incorporate mismatch data into a fault-tolerant learning algorithm \cite{orgenci2001fault}.

\section{Experimental Methods}

\subsection{Data Collection}

Training and testing data was collected from study volunteers in a laboratory setting. All aspects of the study protocol were reviewed and approved by the Dartmouth College Institutional Review Board (Committee for the Protection of Human Subjects-Dartmouth; Protocol Number: 00030005).

\begin{figure}
\begin{center}
\includegraphics[scale=0.22]{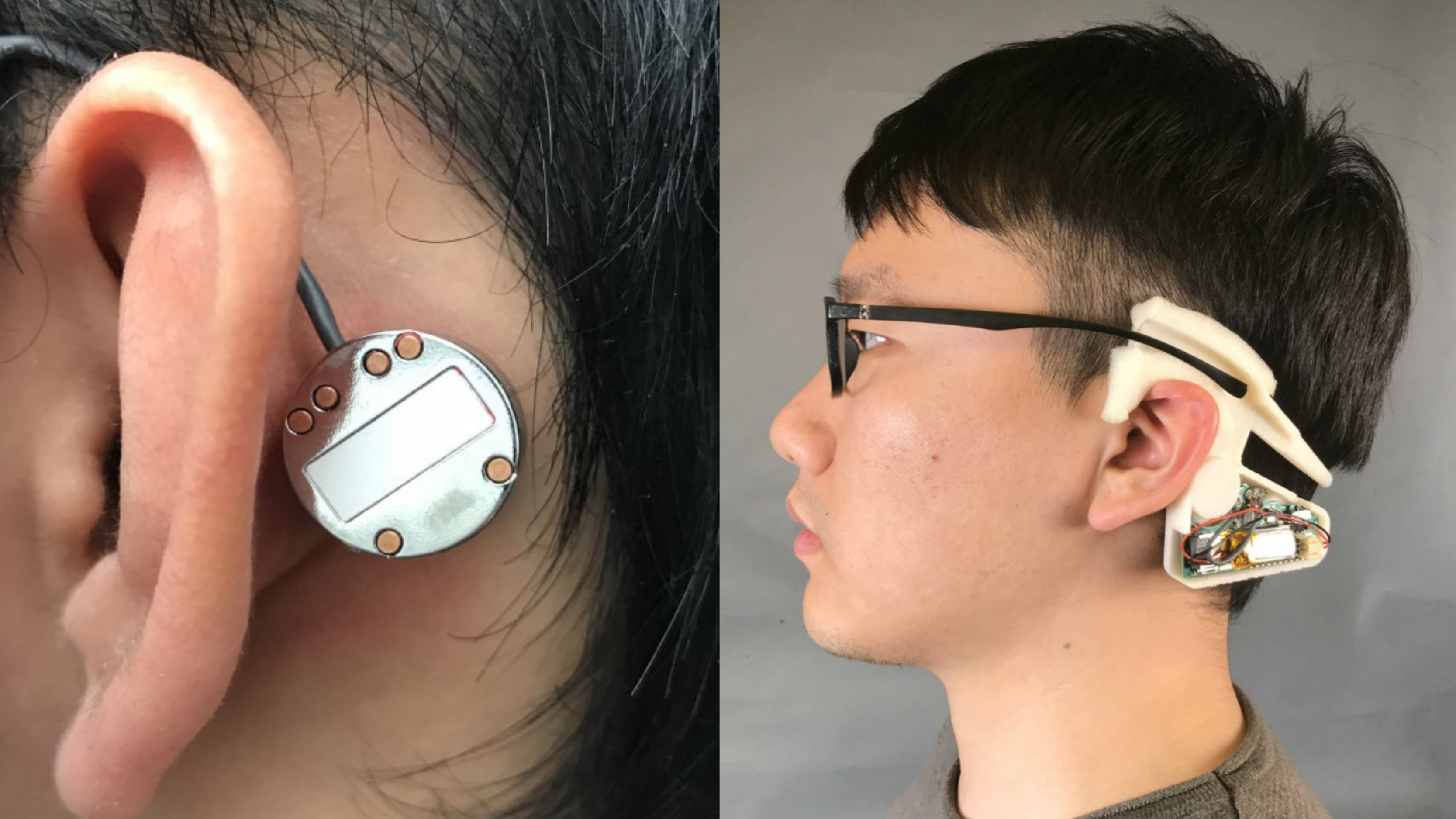}
\end{center}
\vspace{-10pt}
\caption{Left panel: a contact microphone was used to collect acoustic data from the mastoid bone as study participants performed various eating and non-eating tasks \cite{bi2017toward}. Right panel: prototype of the complete wearable device that we are developing for dietary monitoring \cite{bi2018auracle}.}
\label{figure:dataCollection}
\end{figure}

The data used for this study was previously collected in a controlled laboratory setting from 20 participants (8 females, 12 males;
aged 21-30) that were instructed to perform both eating and non-eating-related activities. During these activities, a contact microphone (see \fig{dataCollection}) was secured behind the ear with  a headband, to measure any acoustic signals present at the tip of the mastoid bone \cite{bi2017toward}. The output of the contact microphone was digitized and stored using a 20 kSa/s, 24-bit data acquisition device (DAQ). 

Participants were asked to eat a variety of foods---including carrots, protein bars, crackers, canned fruit, instant food, and yogurt---for at least 2 minutes per food type. This resulted in a 4 hour total eating dataset. Non-eating activities included talking and silence for 5 minutes each and then coughing, laughing, drinking water, sniffling, and deep breathing for 24 seconds each. This resulted in 4 hours total of non-eating data. Each activity occurred separately and was classified based on activity type as eating or non-eating.

We down-sampled the DAQ data to $500$ Hz and applied a high pass filter with a $20$ Hz cutoff frequency to attenuate noise. We segmented the positive class data (chewing), and negative class data (not chewing) into $24$-second windows with no overlap. The positive and negative class data were labelled with the one-hot encoding $(2, 0)$ and $(0, 2)$, respectively. Finally, we extracted the ZCR-RMS and ZCR-ZCR features of the windows to produce $2$-dimensional input vectors to be processed by the AFUA network.

\subsection{Neural Network Training}
\label{section:training}
For training, the AFUA neural network was implemented in Python, using a custom layer defined by the discretized system of \eq{AFUAupdateD}. Chip-specific parameters were extracted for each neuron and incorporated into the custom layers. The AFUA network was trained and validated on the laboratory data (train/valid/test split: $68/12/20$) using the TensorFlow Keras v2.0 package. Training was performed with the \textsc{adam} optimizer \cite{kingma2014adam} and a weighted binary cross-entropy loss function to learn full-precision weights.



Python training was followed by a quantization step that converted the full-precision weights to signed $3$-bit values ($0, \pm1, \pm2, \pm3$). An alternative approach would have been to directly incorporate the quantization process into the network's computational graph \cite{amoh2019optimized}. However, we found that such an approach only slows down training with no improvement in our network's classification performance.

\begin{figure}
\begin{center}
\includegraphics[scale=0.53]{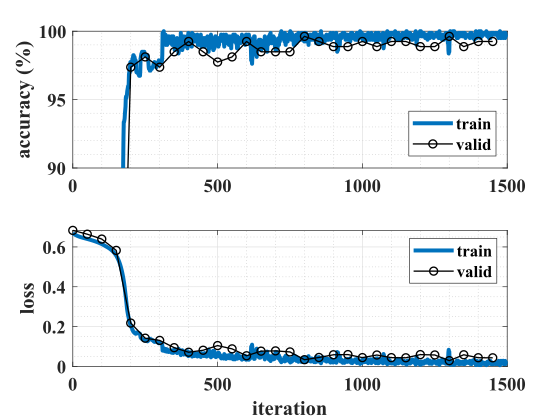}
\end{center}
\vspace{-10pt}
\caption{Accuracy and loss training graphs for discretized AFUA neural network.  We performed training in Python using the TensorFlow Keras v2.0 package. Validation set performance tracked that of the training set, indicating good generalization. The learned weights were quantized and programmed onto the AFUA ASIC's on-chip registers.}
\label{figure:training}
\end{figure}

\subsection{Chip Measurements}

\begin{figure}
\begin{center}
\includegraphics[scale=0.42]{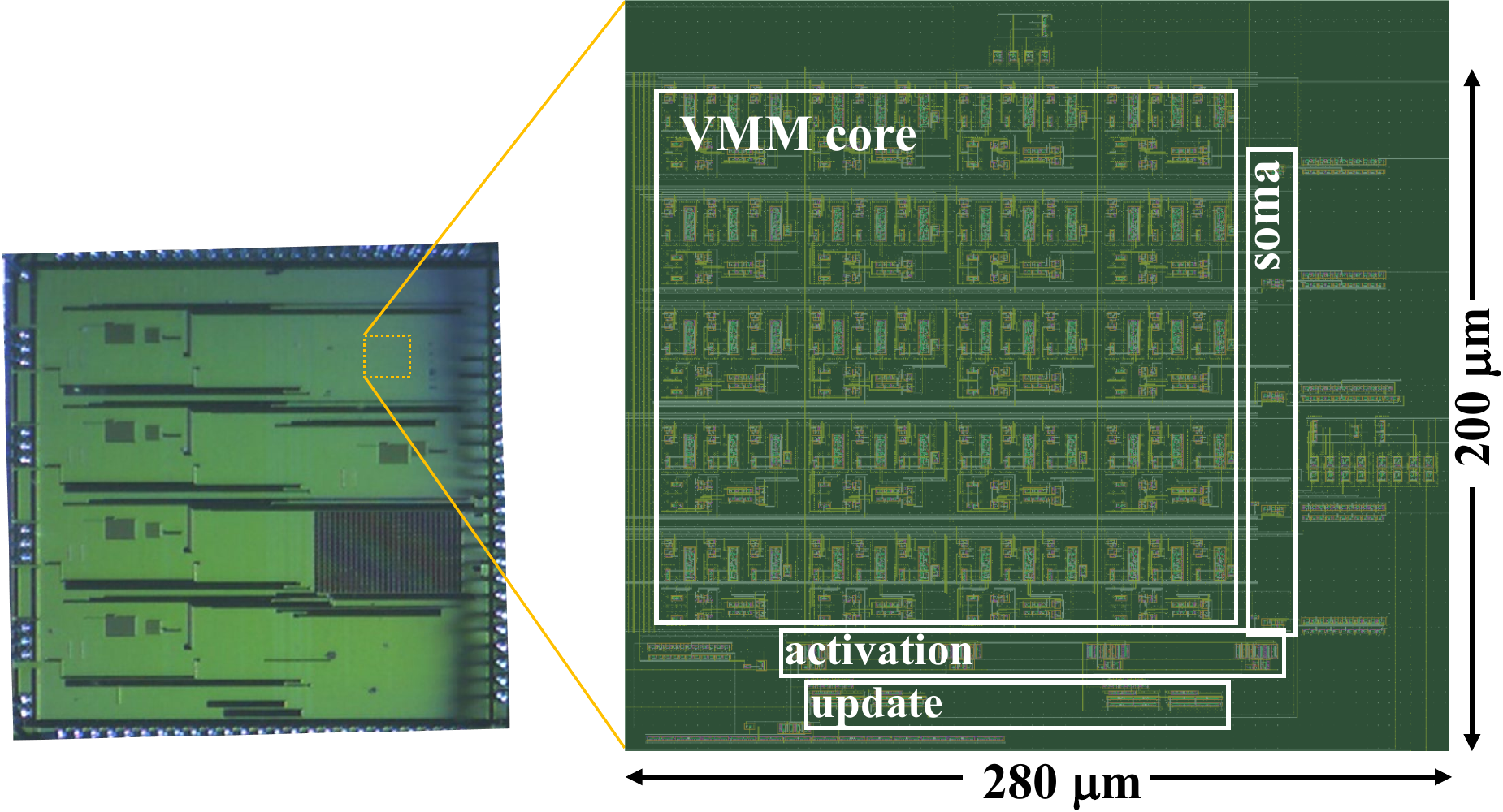}
\end{center}
\vspace{-10pt}
\caption{Die photo of the AFUA ASIC, implemented in a $0.18~\mu$m CMOS process. The synapse circuits (labelled ``VMM core'') consume most of the $200~\mu$m$\times 280~\mu$m circuit area.}
\label{figure:diephoto}
\end{figure}

The AFUA was implemented, fabricated and tested as an integrated circuit in a standard $0.18~\mu$m mixed-signal CMOS process with a $1.8$ V power supply. To simplify the measurement process and associated instrumentation, the ASIC I/O infrastructure includes current buffers that scale input currents by $1/100$ and that multiply output currents by $100$. 

The AFUA neural network was programmed by storing the $3$-bit version of each learned weight onto its corresponding on-chip register in the VMM array. 

The network was then evaluated on the test dataset. Specifically, each $24$-second long window of $2$-dimensional feature vectors from the test dataset was dimensionalized and scaled to $100\times\Iunit$ and input to the ASIC with an arbitrary waveform generator. We set $\Iunit\approx10$ nA with an off-chip resistor. According to \eq{tauIunit}, this \Iunit~creates a time constant of $\tau=0.36~\mu$s, allowing for faster-than-real-time chip measurements---an important consideration, given the large amount of test data to be processed. 

\begin{figure}
\begin{center}
\includegraphics[scale=0.53]{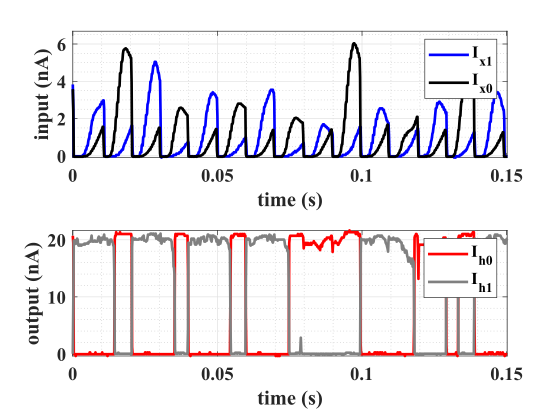}
\end{center}
\vspace{-10pt}
\caption{AFUA chip measurement response to different input patterns ($I_{x1}, I_{x0}$) taken from the test dataset. The circuit's class prediction is encoded as output currents ($I_{h1}, I_{h0}$).}
\label{figure:afuaResponse}
\end{figure}

Output currents $I_{h0}$, $I_{h1}$ were each measured from the voltage drop across an off-chip sense resistor. The ASIC's steady-state response was then taken as the classification decision. An output value of $(I_{h1}, I_{h0}) = (2\Iunit, 0)$ means that the circuit classified the input as eating, while $(I_{h1}, I_{h0}) = (0, 2\Iunit)$ corresponds to non-eating. From these measurements, we calculated the algorithm's test accuracy, loss, precision, recall, and F1-score.

\section{Results and Discussion}
\begin{table*}
    \centering
    \begin{tabular}{|l|c|c|c|c|c|c|} 
    \hline 
        & \textbf{Window Size (s)} & \textbf{Accuracy}   & \textbf{F1-Score} & \textbf{Precision} & \textbf{Recall} & \textbf{Power (mW)}\\
    \hline 
    \hline
        \textbf{This work} & 24  & 0.94 & 0.94 & 0.96 & 0.91 & 0.019\\
    \hline
        \textbf{FitByte \cite{bedri2020fitbyte}} & 5 & - & - & 0.83  & 0.94 & 105\\
    \hline
        \textbf{TinyEats \cite{nyamukuru2020tiny}} & 4 & 0.95 & 0.95 & 0.95  & 0.95 & 40 \\
     \hline
        \textbf{Auracle \cite{bi2017toward}} & 3 & 0.91 & - & 0.95 & 0.87 & \sc{offline} \\
    \hline     
        \textbf{EarBit \cite{bedri2017earbit}} & 35 & 0.90 & 0.91 & 0.87  & 0.96 & \sc{offline} \\
    \hline
        \textbf{AXL \cite{farooq2018accelerometer}} & 20 & - & 0.91 & 0.87  & 0.95 & \sc{offline} \\
     \hline     
    \end{tabular}
    \caption{Comparison between proposed eating detection system and previous solutions. Three of the classification algorithms \cite{bi2017toward, bedri2017earbit, farooq2018accelerometer} were implemented offline; since these are not embedded solutions, their power consumption is not reported.}
    \label{tab:comparision}
\end{table*}

\subsection{Classification Performance}
\figur{afuaResponse} shows the AFUA chip's typical response to input data. The input currents $I_{x1}, I_{x0}$ represent the ZCR-RMS and ZCR-ZCR features extracted from the contact microphone signal. Inputting a stream of $I_{x1}, I_{x0}$ patterns produces output currents $I_{h1}, I_{h0}$, which represent the hidden states of the AFUA network. 

\begin{figure}
\begin{center}
\includegraphics[scale=0.53]{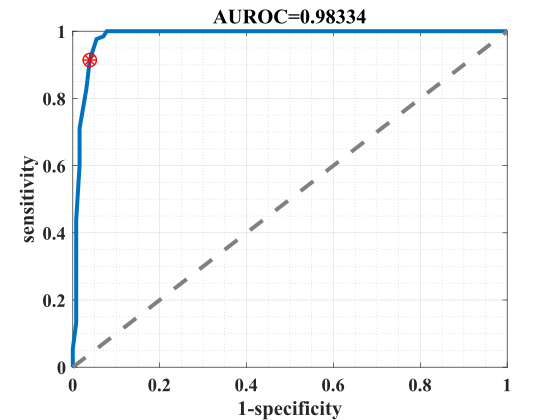}
\end{center}
\caption{Receiver operating curve from AFUA chip measurements. These results were produced from the AFUA chip response to $1.6$ hours of previously-unseen test data. The highlighted point corresponds to a sensitivity of $0.91$ and a specificity of $0.96$.}
\label{figure:ROC}
\end{figure}

According to our encoding scheme,  $(I_{h1}, I_{h0}) = (2\Iunit, 0)$ means that the circuit classified the input as chewing, while $(I_{h1}, I_{h0}) = (0, 2\Iunit)$ corresponds to a prediction of not chewing. But the presence of noise and circuit non-ideality produces some ambiguity in the encoding: some AFUA output patterns can be interpreted as either chewing or not chewing, depending on the choice of threshold used to distinguish between $0~$A and $2\Iunit$. \figur{ROC} is the receiver operating curve (ROC) produced by varying this threshold current. The highlighted point on the ROC is a representative operating point, where the classifier produced a sensitivity of $0.91$ and a specificity of $0.96$. This corresponds to a false alarm rate  of $(1-$specificity$)=0.039$.

\begin{figure}
\begin{center}
\includegraphics[scale=0.53]{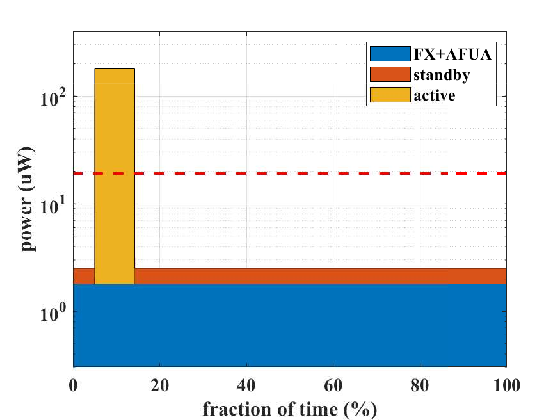}
\end{center}
\vspace{-10pt}
\caption{Power consumption of eating detection system. The feature extraction and AFUA circuitry continuously consume $1.8~\mu$W of power. The microcontroller is active for $9\%$ of the time, during which it consumes $180~\mu$W of power. For the remaining $91\%$ of the time, the microcontoller consumes $0.72~\mu$W while in standby mode. On average (red dashed line), the whole system consumes an estimated $18.8~\mu$W.}
\label{figure:powerAFUA}
\end{figure}

\subsection{System-level Considerations}

In this section, we consider the impact of using the AFUA neural network in a complete eating event detection system. To process a $500$ Hz signal, the ZCR and RMS feature extraction blocks consume a total of $0.68~\mu$W \cite{baker2003micropower}. Also, the AFUA network consumes $1.1~\mu$W, assuming $\Iunit=10~$nA. Finally, a microcontroller from the MSP430x series (Texas Instruments Inc., Dallas, TX) running at $1~$MHz consumes $180~\mu$W when active and $0.72~\mu$W when in standby mode \cite{msp430x}.

The feature extraction and AFUA circuitry are always on, while the microcontroller remains in standby mode until a potential chewing event is detected. The fraction of time the microcontroller is in the active mode depends on how often the user eats, as well as the sensitivity and specificity of the AFUA network. Assuming the user spends $6 \%$ of the day eating \cite{stimpson2020peer}, then, using the classifier operating point highlighted in \fig{ROC}, the fraction of time that the microcontroller is active is
\begin{eqnarray}
\textsc{active} & = & \textsc{eat} \times \textsc{sens} + (1-\textsc{spec}) \times (1-\textsc{eat}) \nonumber \\
& = & 0.06\times0.91 + (1-0.96)\times(1-0.06) \nonumber \\
& = & 0.09.
\end{eqnarray}

So, the microcontroller consumes an average of $180~\mu W\times0.09 + 0.72~\mu W\times(1-0.09)=16.9~\mu$W. As \fig{powerAFUA} shows, the average power consumption of the complete AFUA-based eating detection system is $18.8~\mu$W. If we attempted to implement the system with a front-end ADC (12-bit, 500 Sa/s \cite{bi2018auracle, datasheetCC2640R2F}) followed by a digital LSTM \cite{shin201714, giraldo2018laika}, then the ADC alone would consume over $240~\mu$W of power \cite{datasheetADS1000}. 

Table \ref{tab:comparision} compares our work to other recent eating detection solutions. The different approaches all yield generally
the same level of classification accuracy, but our work
differs in one critical aspect: while others depend on offline processing, or on tens of milliWatts of power to operate, our approach only requires an estimated $18.8~\mu$W. 

\section{Conclusion}
We have introduced the AFUA---an adaptive filter unit for analog long short-term memory---as part of an eating event detection system. Measurement results of the AFUA implemented in a 0.18 $\mu$m CMOS technology showed that it can identify chewing events at a 24-second time resolution with a recall of 91$\%$ and an F1-score of 94$\%$, while consuming $1.1~\mu$W of power. The AFUA precludes the need for an analog-to-digital converter, and it also prevents a downstream microcontroller from unnecessarily processing irrelevant data. If a signal processing system were built around the AFUA for detecting eating episodes (that is, meals and snacks), then the whole system would consume less than $20~\mu$W of power. This opens up the possibility of unobtrusive, batteryless wearable devices that can be used for long-term monitoring of dietary habits.

\section{Acknowledgments}
This work was supported in part by the U.S. National Science Foundation, under award numbers CNS-1565269 and CNS-1835983. The views and conclusions contained in this document are those of the authors and do not necessarily represent the official policies, either expressed or implied, of the sponsors.

\printbibliography
\end{document}